\documentclass[11pt]{article}
\usepackage{graphicx} 
\usepackage{tikz}
\usepackage{hyperref}
\usetikzlibrary{shapes,arrows,positioning}
\usepackage{setspace}

\title{REE-HDSC: Recognizing Extracted Entities \\for the Historical Database Suriname Caribbean}
\author{Erik Tjong Kim Sang\\Netherlands eScience Center}
\date{First version: December 19, 2023\\Current version: \today}

\begin{document}

\maketitle

\begin{abstract}
\noindent
We describe the project REE-HDSC and outline our efforts to improve the quality of named entities extracted automatically from texts generated by hand-written text recognition (HTR) software. We present a six-step processing pipeline, which we tested by processing 19th and 20th century death certificates from the civil registry of the Caribbean island Cura\c{c}ao. We find that the pipeline extracts dates with high precision but that the precision of person name extraction is low. Next we show how name extraction precision can be improved by retraining  HTR models with names, post-processing and by identifying and removing incorrect names. 
\end{abstract}

\begin{spacing}{0}
\tableofcontents 
\end{spacing}

\section{Introduction}
\label{sec-introduction}

REE-HDSC (Recognizing Extracted Entities for the Historical Database Suriname Cura\c{c}ao\footnote{The parent project HDSC has since the start of the project REE-HDSC been renamed to Historical Database Suriname Caribbean})\footnote{REE-HDSC project website: \href{https://research-software-directory.org/projects/ree-hdsc}{https://research-software-directory.org/projects/ree-hdsc}} is a project of the Radboud University Nijmegen and the Netherlands eScience Center aiming at pushing forward the state-of-the-art of information extraction (IE) from data created by automatic  hand-written text recognition (HTR). As a data domain, we have chosen for death certificates of the period 1831 to 1950 from the Caribbean island Cura\c{c}ao, a former colony of The Netherlands. In the parent project HDSC, information extraction from digital scans of the death certificates is performed by human volunteers. REE-HDSC aims to investigate if the combination of HTR and IE can speed up this process. 

\begin{figure}[b]
\begin{center}
\begin{tikzpicture}[
    block/.style={rectangle, draw, text width=3.5em, text centered, minimum height=4em},
    line/.style={draw, -latex'}
]
\node [block, text width=4em] (layout) {Layout\\Analysis};
\node [block, right=0.6cm of layout, text width=4.5em] (baseline) {Baseline\\Detection};
\node [block, right=0.6cm of baseline, text width=7em] (htr) {Hand-written\\Text Detection};
\node [block, right=0.6cm of htr, text width=5.5em] (ner) {Entity\\Recognition};
\node [block, right=0.6cm of ner, text width=5em] (nc) {Name Correction};
\node [block, right=0.6cm of nc] (el) {Entity\\Linking};
\path [line] (layout) -- (baseline);
\path [line] (baseline) -- (htr);
\path [line] (htr) -- (ner);
\path [line] (ner) -- (nc);
\path [line] (nc) -- (el);
\end{tikzpicture}
\end{center}
\caption{Flow diagram of automatic death certificate analysis process with six different tasks}
\label{fig-pipeline}
\end{figure}
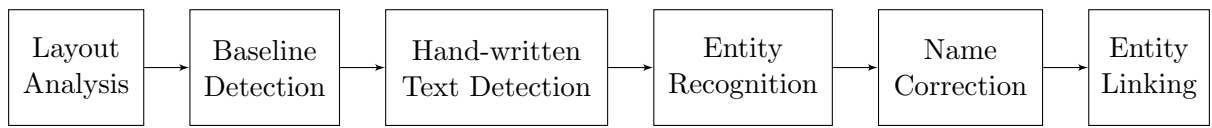

We suggest to divide the automatic death certificate analysis process in six\footnote{We are considering adding a seventh task: reordering the lines in the HTR output, after hand-written text recognition and before entity recognition.} tasks (see Figure \ref{fig-pipeline}):

\begin{enumerate}
    \item Layout analysis: determine the layout of the certificates: how many columns of text do they contain?
    \item Baseline detection: find the locations of the lines of texts
    \item Hand-written text detection: recognize the text of the certificates. The certificates contain both printed text and hand-written text
    \item Entity recognition: identify the interesting entities in the texts: person names, locations, dates, times, ages and professions
    \item Name correction: combine name parts from different locations on certificates
    \item Entity linking: identify identical person names in different certificates and link them
\end{enumerate}

Layout detection, baseline detection and hand-written text detection are performed by the program Transkribus\footnote{Transkribus website: \url{https://readcoop.eu/transkribus}}. For entity recognition, name correction and entity linking, we will explore using regular expressions \cite{hoek2023}, machine learning and ChatGPT\footnote{ChatGPT website: \url{https://chat.openai.com}}.

The output of the automatic tasks needs to be checked and corrected. This involves manual labour. However, we also want to perform an automatic quality assessment for all six tasks. At any point of the analysis process, analyzed documents of poor quality need to be identified and to be put aside for manual inspection and correction.

\section{Data preparation}

The data consist of tens of thousands of scans of death certificates. Before processing them, we first inspect the data, remove duplicates and non-certificate scans and make an overview of how much data is available for each time period. 

\subsection{Data analysis}
\label{sec-data}

The data consists of scans of death certificates from the civil service of the island Cura\c{c}ao for the years 1831-1950. The scans are available in jpeg format. Some scans are also available as pdfs. Each year is a separate folder in the data set. Each folder contains several subfolders per district. The district folders contain the scans. If there are pdfs available, these are stored in district folders in a separate pdf folder in the year folder (years 1831-1840, 1848-1851 and  1934). 

The number of districts varies over the years: nine districts (1831-1863), five districts (1864-1924) and three districts (1925-1950). Surprisingly the year 1852 also contains scans for an unexpected tenth district while 1901 and 1912 have a folder for a sixth district. The extra district data for 1912 is for the fifth district but for the other two years the source is unknown. The districts are conveniently named 1 to 9, where district 1 is usually named {\it Stad} (city) corresponding to the capital Willemstad, where more than half of the population of the island lives. The other districts are called {\it Buiten} (outer) districts.

\begin{figure}[t]
\begin{center}
\includegraphics[width=150pt]{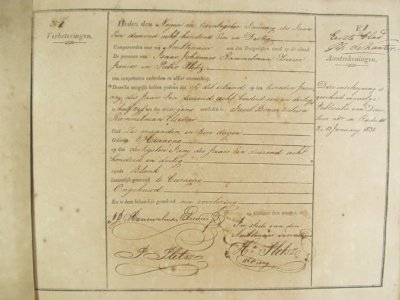}
\includegraphics[width=150pt]{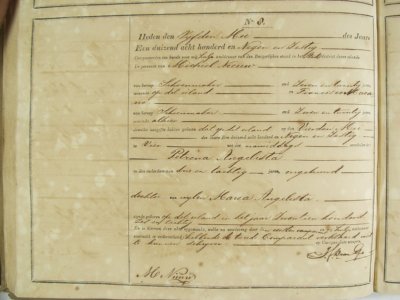}
\includegraphics[width=150pt]{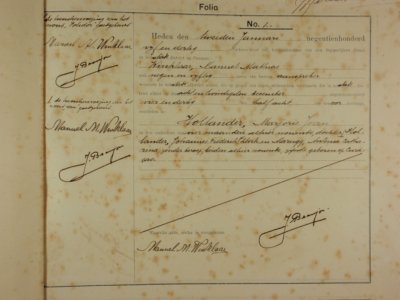}
\end{center}
\caption{Three main certificate form formats in the data: three-column (1831-1869, left), early two-column (1869-1934, middle) and late two-column (1935-1950, right). The main text of the forms can be found in the widest center column. However, narrower margin columns may contain hand-written notes which need to be processed as well. }
\label{fig-form-formats}
\end{figure}

The year files (version data 22 August 2022) contain 77,352 jpeg files. Not all of these files are scans of death certificates and some of the scans may be duplicates. Among the death certificates, we identified three main formats of forms: three-column (1831-1869\footnote{The switch between three-column format and early two-column format was made on 1 May 1869}), early two-column (1869\footnotemark[\value{footnote}]-1934) and late two-column (1935-1950), see Figure \ref{fig-form-formats}. The main difference between the two two-column formats is the use of a more modern spelling in the late version, like "verscheen" instead of "compareerden" while removing words like "des", "dezes, "dewelke" and "jaars". The main text of the forms can be found in the widest central column. However, narrower margin columns may contain hand-written notes and these need to be processed as well.

The year folders are named {\tt O.R. 1111}, presumably from {\it Overlijdens Register} (death register) where {\tt 1111} is a four-digit year number ranging from 1831 to 1950. One folder was missing: {\tt O.R. 1911} which proved to be a subfolder of {\tt O.R. 1914}. Scans from the city registry can be found for each year. Not all outer districts are represented in each year, see appendix \ref{sec-missing-districts}, perhaps because no deaths were registered for those districts in those years. However, for 1870 all outer districts are missing which seems like a case of missing data\footnote{The later to be discussed extra Excel annotation file contains data on about 130 certificates from the outer districts from the year 1870 not included in our scans.}.

The district folder names have the format {\tt O.R. 1111 Stad} or {\tt O.R. 1111 Buiten 2e distr} where {\tt 1111} is a four-digit year number ranging from 1831 to 1950 and {\tt 2} is a number ranging from 2 to 10. In the first 10 years of the collection (1831-1840), the districts were divided in three groups: Midden (middle), Oost (east) and West, and the folder name format was  {\tt O.R. 1111 Buiten Group 2e distr}, where {\tt Group} was the group name. In the 1869 five-district arrangement, districts were numbered from east to west \cite{renkema1981}. If that is also the case for the 1841 arrangement, the districts correspond to the numbers 2--9 as: Oost 1, Oost 2, Oost 3, Midden 1, Midden 2, West 3, West 2 and West 1\footnote{In the subfolder {\it renamed} for the year 1848, West 1 and West 3 were reversed in the order.}.

The file names of the scans have the format {\tt O.R. 1111 District 333.JPG} where {\tt 1111} is a four-digit year number ranging from 1831 to 1950, {\tt District} is the district name and {\tt 333} is a three-digit file (certificate) number, which corresponds with the folio number written in the document text. It is useful that the year and the district are mentioned in the file names because this makes it possible to put files of different years in the same folder without name clashes. Some file names have a single letter behind the numbers. This indicates the presence of additional documents like letters or notes accompanying the death certificate.

\begin{figure}[t]
\begin{center}
\includegraphics[width=230pt]{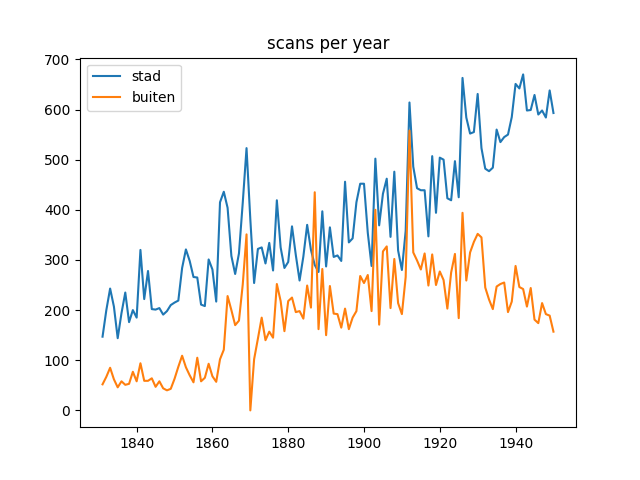}
\includegraphics[width=230pt]{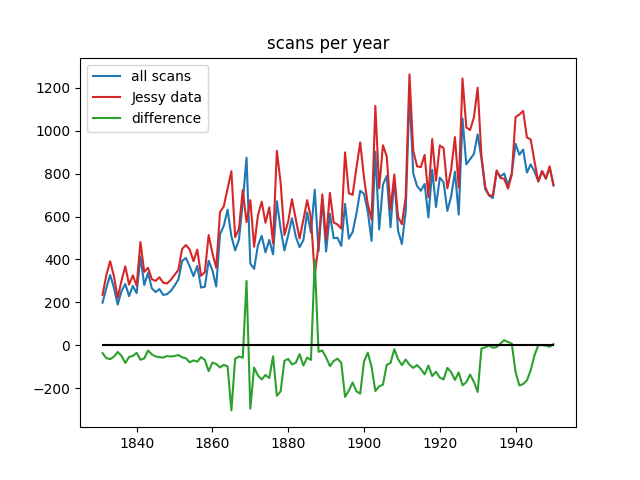}
\caption{Number of death certificates in the collection per year after data cleaning. Left graph: there are two groups of certificates: certificates from the capital (stad) and certificates from the other districts (buiten). Right graph: interestingly the extra Excel annotation file contains data for more certificates than we have scans, with the years 1869 and 1887 as notable exceptions. These years may contain several duplicates in our collection.}
\label{fig-scans-per-year}
\end{center}
\end{figure}

The 1910 year directory contains an extra Excel file with 60.000+ names of deceased people in the death certificates with their death days\footnote{In the thesis of Hoek \cite{hoek2023} this file is called {\it death register database} and is discussed in its chapter 6}\footnote{The data from this extra Excel file are already shared by the Archive of Cura\c cao: https://www.nationaalarchief.cw/api/picturae/genealogie/persons?f={\%}7B{\%}22search{\_}s{\_}deed{\_}type{\_}title{\%}22:{\%}7B{\%}22 v{\%}22:{\%}22Overlijdensakte{\%}22{\%}7D{\%}7D}. It seems that someone has already extracted interesting information from most of the death certificates. There are different versions of the excel file, we use {\tt Overlijdensmerged.csv}. This file contains data for more certificates than we have scans, see Figure \ref{fig-scans-per-year}. 

\subsection{Data cleaning}

The data consist of 77,352 scan of death certificates of Cura\c{c}ao. We found two problems in the data. First, several scans did not contain a death certificate but a note. This note could consist of a small piece of paper or letters of several pages. The notes are usually identifiable by their name: instead of a numeric note number in the file name, their names consist of a number plus a character. We have verified that all note names refer to the relevant certificate number. Where a note accompanied a certificate we appended a character (a, b, c, ...) to the certificate number in the file name to indicate the link.

The second problem was that there were several duplicate scans in the data. We found two kind of duplicates: duplicate files, where the complete contents of two files was identical, and duplicate scans, where a certificate was scanned twice. Duplicates copies of a file have been removed from the data set while duplicate scans have been put in a separate folder (x-duplicates). The latter could be useful in cases where a scan is difficult to read. Some of the duplicate scans involve the same piece of paper being scanned twice. Others involve archive duplicates being scanned as well as the originals.

The data cleaning work also involved enforcing the same naming scheme in all the data folders. Strict enforcement of the naming schemes was a major way for detecting duplicates. Duplicate detection and duplicate removal decreased the size data of the data set from 77,352 scans to 68,520 scans. The data cleaning process was stopped on 15 May 2023 and later finished by Lisa Hoek, who distributed a new version of the data set on Google Drive on 12 December 2023.

\section{Pipeline implementation}

In this section we describe the six different parts of the information processing pipeline presented in the Introduction section.

\subsection{Layout analysis}

Since the text on the scans is written in more than one column, we need layout analysis to find the positions of the columns. Without layout analysis, sentences written on the same line in different columns will be treated as one sentence.

\begin{figure}[t]
\begin{center}
\includegraphics[width=230pt]{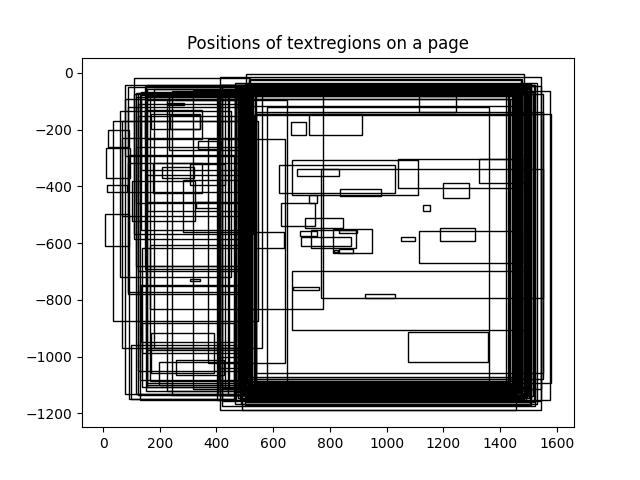}
\includegraphics[width=230pt]{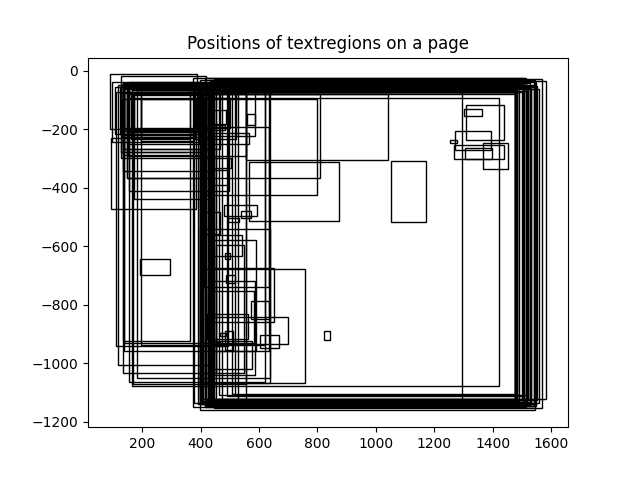}
\caption{Text regions identified in two-column texts (1870-1950) and in three-column texts (1831-1869, right) by the model P2PaLA\_Curacao \_bestModel from Hoek \cite{hoek2023}. The recognition of the main text column seems to work fine for two-column text but not for three-column text, where the center text column has been combined with the right margin column in 97\% of the data.}
\label{fig-layout-detection}
\end{center}
\end{figure}

We ran a layout analysis with the best layout detection model trained by Hoek \cite{hoek2023}, P2PaLA\_Curacao \_bestModel\footnote{Time stamp of the model: 2023-03-18 17:14:47.387854. The model can be found in our Transkribus environment under: Tools / P2PaLA structure analysis tool / Select a model for recognition},  on a selection of files, the first city certificates of each year (1831-1950). Note that P2PaLA is not the standard layout algorithm of Transkribus but an alternative to the default. The model was trained on two-column texts. We found that the model was unable to predict the layout of three-column text, see Figure \ref{fig-layout-detection}. The right margin column was combined with the main center column in accordance with the two-column style training data. In the Figure, we have converted each identified text region to the minimal-sized rectangle that included it.

\begin{figure}[t]
\begin{center}
\includegraphics[width=230pt]{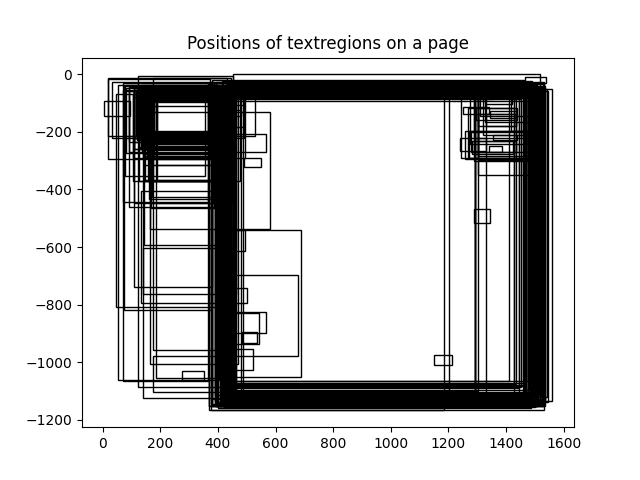}
\includegraphics[width=230pt]{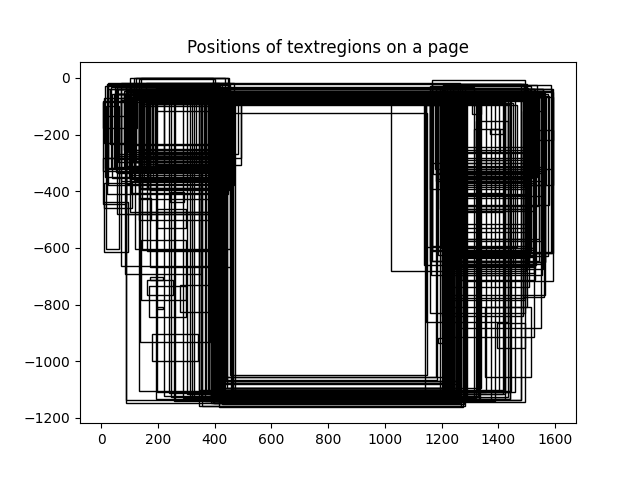}
\caption{Text regions identified in 100 randomly selected three-column texts (1831-1868) with a two-column layout model (left) and with a three-column layout model (right). The three-column model more often predicts a center column with space for right margin text (75\% vs 6\%).}
\label{fig-layout-detection-improved}
\end{center}
\end{figure}

In order to overcome the problem of the three-columns text layout, we trained an additional P2PaLA model on only three-columns text. For this purpose we used the first deceased documents for the city district for each of the years 1831-1868. We tested the new model three-columns(6)\footnote{Time stamp of the model 2023-04-24 14:52:05.760327. The model can be found in our Transkribus environment under: Tools / P2PaLA structure analysis tool / Select a model for recognition} on 100 other documents from the years 1831-1868. Figure \ref{fig-layout-detection-improved} displays the results in comparison with an analysis with the two-column layout model. Unsurprisingly, the three-column model works better for processing three-column layout texts. Hence this model was used in all later experiments with data from the three-column period (1831-1868) for detecting text layout (and finding baselines).

Instructions used for running and training the layout models in Transkribus can be found in the appendix. Both layout models described in this section also perform the baseline detection task described in the next section. After running layout detection, the results can be inspected and corrected in Transkribus, see the software's website\footnote{Manually correcting layout analysis with Transkribus: https://readcoop.eu/transkribus/howto/how-to-transcribe-documents-with-transkribus-introduction/} for instructions

\subsection{Baseline detection}

Baseline detection in the program Transkribus involves finding the positions of lines of text in a scanned document without considering the shapes of the characters \cite{readcoop2021}. These are different from line regions, which include the shape of the characters. This task is included in the layout analysis task described in the previous section. No extra actions need to be executed to perform this task. No evaluation was performed for this task. It would have been interesting to check how often lines with big white space chunks get broken up in smaller parts. 

\subsection{Hand-written Text Detection}

Hand-written Text Detection is the final step of the automatic text recognition process, after layout detection and baseline detection. We use the model by Hoek \cite{hoek2023}: HTR\_Curacao\_bestModel\footnote{Time stamp of the model: 07-03-23. The model can be found in our Transkribus environment under: Tools / View models / text}. Before the model was run, we set one parameter: "Language Model" to the value: "Language model from training data". The language model is used when choosing between alternative options for recognized words. By using the language model from the training data, Transkribus will have a better chance of finding correct words and names. 

\subsection{Entity Recognition}

For entity recognition testing, we reused a sample of 100 randomly chosen scans for layout detection: three-columns-100 (1831-1868). We applied the earlier mentioned models for (three-column) layout detection, baseline detection and hand-written text recognition to the files. Next we applied a model for Dutch named entity detection (wietsedv/bert-base-dutch-cased-finetuned-udlassy-ner) \cite{devries2020} to the texts combined with manually written regular expressions for extracting names of deceased persons and death dates from the documents. The regular expressions were developed based on two other data sets (Training\_set\_V2 and Sample\_regex)\footnote{The regular expressions for entity detection by Hoek \cite{hoek2023} were not yet available at the time of this experiment.}. The test results were compared with a gold standard results which accompanied the original data (see Table \ref{tab-entity-recognition}).

\begin{table}
    \begin{center}
        \begin{tabular}{l|ccc|ccc}
                         & \multicolumn{3}{c|}{\bf Deceased names} & \multicolumn{3}{c}{\bf Death dates}\\
            {\bf Method} & Found & Exact & Partial & Found & Exact & Corrected \\\hline
            Regular expressions &  95\% & 17\% & 55\% &  76\% & 37\% & 58\% \\
            GPT3.5              & 100\% & 33\% & 83\% & 100\% & 82\% & 90\% \\
            GPT4                & 100\% & 34\% & 81\% & 100\% & 84\% & 91\% \\
        \end{tabular}
    \end{center}
    \caption{Evaluation of entity recognition. ChatGPT outperforms regular expressions both for extracting deceased names and death dates. We evaluated the percentages of found entities, exactly correct entities, partially correct entities (Levenshtein distance up to 3) and correctness after post-correcting obviously incorrect death years.}
    \label{tab-entity-recognition}
\end{table}

This approach found deceased names for 95\% of the documents. However for only 17\%, the names were completely correct. When names up to a Levenshtein distance of three were considered correct, the accuracy  moved up to 55\%. Some of the mistakes involved certificates mentioning only the first names of the deceased while the gold data included the surname of the mother, which was mentioned somewhere else in the death certificate. The automatic analysis found a death date for 76\% of the documents and 37\% of them were correct. The correct year of the documents is encoded in the file name. When the incorrect years were automatically changed to these years, the death date accuracy moved up to 58\%. The analysis may generate several candidate death days per certificate. The regular expressions perform worse than reported by Lisa Hoek \cite{hoek2023}. Unlike us, Hoek allows fuzzy matching but this did not improve performances. Most likely, the difference is caused by us using the Excel file accompanying the data as golden standard while Hoek used a separately developed golden standard which is more accurate \cite{hoek2024}.

We repeated this evaluation, this time using ChatGPT (GPT4 and GPT3.5)\footnote{The online version of ChatGPT3.5 is free. We paid a subscription (20 USD per month) to be able to test the online ChatGPT4 version. The ChatGPT3.5 API costs 0.07 cents per request.}\footnote{We only used certificates of more than 150 years ago (1831-1868) for evaluating ChatGPT.} \cite{bubeck2023sparks} for extracting the information. The system extracted deceased names and death dates for all documents. GPT4 got 34\% of the names completely correct and allowing a maximum Levenshtein distance of three raised the accuracy to 81\%. GPT4 correctly extracted 84\% of the death dates. After automatically correcting obviously incorrect years, the accuracy rose to 91\%. GPT3.5 performed at the same level: 33\%/83\% for deceased names and 82\%/90\% for deceased names. We discovered that GPT3.5 had problems processing more than one document at a time\footnote{Processing batches of 11 documents reached a death date accuracy of 43\% while batches of size one performed at 82\% correct.}. Therefore all experiments with the GPT models involved using one document and one information request per question. The improvement of ChatGPT over the regular expressions is much larger for dates than for names because dates use a limited vocabulary which enables the system to detect and correct character errors.

In the names found by ChatGPT3.5, there are words from which we know that they are names (like {\it Maria}) and words which we do not know as names (like {\it Mariak}). Replacing the unknown words in suggested names identified by the closest and most frequent known name words led to a modest improvement of 6\%. We used names of later years (1869-1950) as source of the known name tokens. The known name tokens were able to signal problems with the identified names: 78\% incorrect names contained unknown tokens while only 14\% of the correctly recognized names had an unknown token. Allowing any order of the names as correct increased led to one extra correct name. The same was true for appending the surname of the mother to names without surnames, if the mother's name was known. The fact that ChatGPT version 3.5 and version 4 perform similarly is good news because at the moment only version 3.5 allows large-scale access via a software API.

\subsection{Name correction}
\label{sec-name-correction}

The names of the deceased on the certificates can be incomplete or incorrect. A frequently occurring example of an incomplete name consists of only the first name(s). In that case the complete name of the deceased person requires adding the surname of the father or the mother when these are mentioned on the certificate. At other times the name of the deceased is corrected in a margin note. In those case the margin note names are correct.

With entity recognition, we find all names mentioned the certificate. For entity linking it is crucial that we find the right name of the deceased person. Our current strategy is to find and keep all names from the certificate that refer to the deceased person. This may require combination of different names mentioned on the certificate, which is what the current name correction process is about. Presently we do not have a software solution for this module.

\subsection{Entity Linking}
\label{sec-entity-linking}

For entity linking, we used the data file which came with the scans, {\tt Overlijdensmerged.csv}, which partially covers the death certificates of Cura\c{c}ao for 1831-1950. The file contains data of 78,564 certificates, some of which are duplicates. The certificates include data of 129,430 persons, mainly of deceased persons but also of their parents, spouses and the people that reported the deaths. This makes it possible to link different names in the certificates.

The primary information for linking persons are their names. However, the certificates contain many duplicate names which do not refer to the same person, as can be seen for the name Esther Curiel, for which we have nine death certificates from nine different years. At least one other fact is required for linking the names. We used the birth year, which is the next most frequent fact related to persons in the data. Often the birth year needs to be derived from the reported age. 

When we look for link information based on birth year and exact names, we find useful personal data for 7,887 unique names. For these names, 7,210 unique names corresponded with people that had reported deaths. We expected these witnesses to appear as deceased persons as well but we only found a unique death date for 427 of the names (6\%). Table \ref{tab-entity-linking} contains two examples of link groups. In 106 of the cases, the linked information specified "father" as a role. The results of this linking are lower than expected which could indicate that the data set is incomplete, or that the names in the records are not spelled consistently, or that the mobility of the inhabitants of Cura\c{c}ao is much higher than expected, that is that many died somewhere else.  

\begin{table}
\begin{center}
\begin{tabular}{l|lccll}
{\bf Name} & {\bf Role} & {\bf Year} & {\bf Age} & {\bf Profession} & {\bf Other} \\\hline 
{\bf Johan Frederik Garmers} & witness   & 1873 & 31 & shoemaker & \\
                             & father    & 1873 & 31 & shoemaker & Maria Nicolina Garmers \\
                             & deceased  & 1901 & 59 & smith     & 07-12-1841\\\hline
{\bf Louis Martis}           & informant & 1874 & 37 & farmer    & \\
                             & deceased  & 1878 & 41 & farmer    & 1837 \\
                             & deceased  & 1913 & 75 & farmer    & 1837 \\
\end{tabular}
\end{center}
\caption{Example of a correct entity link group (top) and an incorrect link group (bottom)}
\label{tab-entity-linking}
\end{table}

\section{Improving name recognition quality}

The names of the deceased are only identified with an accuracy of about 33\% (Table \ref{tab-entity-recognition}). There are two reasons for this. First, the names on the scans can be different from the names in the gold standard data. We discussed this issue in section \ref{sec-name-correction}. The second reason is the automatic hand-written text recognition (HTR), which produces many character errors\footnote{It is difficult to determine the exact number of character errors in names. Finding this number requires knowing exactly where the names are in the text and which characters in the Transkribus output correspond with characters in the gold standard. The latter is a hard task because Transkribus may generate a different number of characters than in the text and it may move around phrases.}. We can deal with these errors in two ways: first by creating more training data so that we can build better HTR models and second by post-processing the output of the models. Additionally, we can also develop tests for estimating whether generated names are correct or wrong and only use the correct names found by the system.

\subsection{Providing more name examples to HTR}
\label{sec-processing-name-snippets}

We generated more training data by extracting the image parts of the names of the deceased persons from scans from three decades: 1831-1839, 1840-1849 and 1920-1929. The correct associated names had already been identified by volunteers. However, not all the names in the metadata were the same as those on the scans. Some contained extra or different surnames from other places of the certificate and others contained no name, for example for stillborn children. We discarded all non-matching names and only kept scan parts which contained names matching the metadata.

We found out that the baseline detection in Transkribus did not work very well for isolated text snippets. So we combined multiple image snippets on a single text page to get the same relative character sizes on a page. Twelve lines of snippets fit on a scan and for evaluation purposes we only used one name snippet per line. The pages were processed with the P2PaLA layout model {\it combined images (7513) one name per line (first 100 pages)} (ID=56916), specially developed for this task, and these results were manually corrected.

\begin{table}
\begin{center}
\begin{tabular}{c|c|ccc}
& {\bf 3 decades} & {\bf 1830-1839} & {\bf 1840-1849} & {\bf 1920-1929} \\\hline
before post-processing & 60\% & 68\% & 69\% & 41\% \\
after post-processing  & 60\% & 68\% & 66\% & 43\% \\
quality estimation     & 85\% (63\%) & 93\% (65\%) & 88\% (67\%) & 73\% (44\%) \\
\end{tabular}
\end{center}
\caption{Performance on recognizing names of deceased persons on certificates from Cura{\,c}ao of Transkribus models retrained specifically on names. The percentages indicate accuracies. The percentages on the bottom row are only relevant for a part of the data (sizes indicated between brackets)}
\label{tab-htr-improvement}
\end{table}

Next, the 478 scans were used to train a new model based on our model {\it HTR\_Curacao\_best\-Model} (ID=50568). 431 scans (90\%) were used for training and 47 scans (10\%) for validation. After 250 epochs of training the new model achieved a character error rate of 2.0\% on the train data and 5.5\% on the validation data. The name accuracy on the validation data proved to be 60\% when using the train data as language model (see Table \ref{tab-htr-improvement}), which is an improvement compared to the 33\% we started with but is yet far away for the 90\% we are aiming at.

We tested an additional post-processing process which replaced each unknown word in the HTR output with the closest word with a frequency of at least 2. Ties were resolved by selecting the word with the largest frequency. For this process we used a list of 147,338 names\footnote{File name: Namen en beroepen Curacao.csv} obtained from Bj\"orn Quanjer. Post-processing did not improve performance: the accuracy for the three decades remained 60\% (see Table \ref{tab-htr-improvement}). We suspect that the language model used by the HTR processing already does a good job on avoiding non-words. One reason for this could be that a variant of the correction step we implemented here is already part of Transkribus. 

Finally we tested if we could predict which words in the HTR output were correct and which were wrong. For this task we accepted all words which occurred in our name list and rejected all other words. This rule performed reasonably well: it got 85\% of the names correct with a coverage of 63\% (Table \ref{tab-htr-improvement}). This means that we  can identify 63\% of the names with an accuracy of 85\%. The remaining 37\% of the names would have to be processed by humans. This process worked best for the decades 1830-1839 and 1840-1849, where it identified 66\% of the names with an accuracy of around 90\% (Table \ref{tab-htr-improvement}).

Although this experiment reached satisfying performances, it has a few limitations. First, the results were achieved for scans with manually corrected layout. Automatic layout detection will introduce errors which will degrade the quality of of the recognized names. And second, we only evaluated the quality of names. The quality of the rest of the text of the certificates has not been tested. It is quite possible that the improvement of the name quality has degraded the quality of other parts of the certificate texts.

\subsection{Applying other HTR software: Loghi}
\label{sec-loghi}

The Huygens Institute is developing the Loghi\footnote{Loghi software website: https://github.com/knaw-huc/loghi} HTR software for processing millions of pages of 17th and 18th century Dutch text of Dutch East India Company (VOC) in the Globalise project\footnote{Globalise project website: https://globalise.huygens.knaw.nl/}. Although the software is trained for different centuries, it is still worthwhile to evaluate its performance on our data. Loghi shares with Transkribus that its team is responsive to suggested improvements. Loghi has as an advantage that HTR is run locally so there are no waiting queues with jobs of other users and no paid credits are required to process texts. 

We have installed Loghi with the proposed models "general" for laypa and "generic-2023-02-15" for loghi-htr adn tested Loghi with one document of our collection: O.R. 1887 Stad 411.JPG. Nonadapted Loghi obtained a character error rate of 22\% on this document which is much higher than Transkribus: 6\% after training on our data. The biggest problem was text region detection: Loghi found seven regions in our document rather than two and it was unable to put them in the right order. We evaluated three other region models provided by Loghi: cBAD, republic and republicprint but these performed even worse, see Table \ref{tab-loghi-region}.

We attempted to retrain the region detection models of Loghi but this was unsuccessful: our hardware (16Gb memory) proved to have insufficient memory to run Loghi's region training software Laypa. Retraining the hand-written text recognition did work, it took about 15 hours on a laptop without a GPU.

\begin{table}[t]
\begin{center}
    \begin{tabular}{lccc}
        {\bf Region model} & {\bf HTR model} & {\bf Number of regions found} & {\bf CER} \\\hline
        general baseline       & generic-2023-02-15 &  7 & 22\% \\
        cBAD baseline          & generic-2023-02-15 &  8 & 59\% \\
        republic baseline      & generic-2023-02-15 & 11 & 38\% \\
        republicprint baseline & generic-2023-02-15 & 42 & 62\% \\
    \end{tabular}
    \caption{Performance of four default Loghi region models on our file O.R. 1887 Stad 411.JPG measured by character error rate (CER).
    The incorrect proposed orders of the regions have been adjusted manually before measuring CER.}
    \label{tab-loghi-region}
\end{center}
\end{table}

\subsection{Retraining HTR language models}

Apart from retraining HTR models with extra name data, we also tried to improve name recognition by adding a list with 140,601 Dutch colonial names to the language model of a trained HTR model. Language models are character-based models which can optionally be used in Transkribus to choose between alternatives suggested by the hand-written text recognition process.  Retraining these models with extra data is not a standard option in Transkribus so we needed help from the Transkribus team (Sara Mansutti and Sebastian) to make this possible. The performance of the model can be found in Table \ref{tab-retrained}. Most importantly the model with the name list (HTR\_Curacao\_bestModel+) does not perform better on recognizing names than a general model trained on more data (Curacao\_Dutchess)\footnote{HTR\_Curacao\_bestModel was trained on 99 death certificates while Curacao\_Dutchess was trained on 278 death certificates.}. So for improving the name accuracy it is better to train the model on more general data.

\begin{table}[t]
     \begin{center}
    \begin{tabular}{rccc}
               {\bf Platform} & Transkribus              & Transkribus       & Loghi \\
             {\bf Model Name} & HTR\_Curacao\_bestModel+ & Curacao\_Dutchess & generic \\
              {\bf Model ID } & 50568                    & 55666             & 2023-02-15 \\
     {\bf Reported model CER} & 6.0\%                    & 3.7\%             & unknown \\\hline
           {\bf Measured CER} & 5.3\%                    & 3.9\%             & 35.8\% \\
   {\bf CER after correction} & 5.2\%                    & 4.0\%             & 35.1\% \\
          {\bf Name accuracy} & 70\%                     & 75\%              & 55\% \\
    {\bf Profession accuracy} & 78\%                     & 80\%              & 52\% \\
    \end{tabular}
    \end{center}
    \caption{Performances of HTR models on the test collection Sample\_test\_1 with 100 death certificates from Cura\c cao. Loghi was not retrained on our data. HTR\_Curacao\_bestModel+ uses a language model which was retrained with 140,601 additional Dutch colonial names. The row CER after correction refers to data where the printed text was corrected to match the known format of the certificates. Name and profession accuracy includes identified names that were within a character distance of 3. The average length of a name was 19 characters and the average length of a profession was 11 characters.}
    \label{tab-retrained}
\end{table}

\section{Related work}

We are not the first to apply hand-written text recognition to archival manuscripts. Nockels et al. \cite{nockels2022} examined 381 publications which describe using Transkribus for recognizing hand-written text. They mention that Transkribus is the most commonly used tool for processing hand-written text while Tesseract\footnote{Tesseract website: https://github.com/tesseract-ocr/tesseract} is often employed for recognizing printed text, a task which is known as optical character recognition. The cited publications include Gr\"{u}ning et al. \cite{gruning2018}, which discusses accurate identification of baselines, a topic very relevant for our study. Two of the other cited works are quite extensive: Milioni \cite{milioni2020}, which discusses the use of Transkribus and presents a summary of experiences of users, while Muehlberger et al. \cite{muehlberger2019} offers applications and limitations of Transkribus in hand-written text recognition for historical manuscripts, in particular mentioning that low quality HTR results (CER$>$10\%) can still successfully be used for searching in.

The thesis of our colleague Lisa Hoek \cite{hoek2023} has the same focus as this report: investigate the application of automatic hand-written text recognition in the project HDSC (Historical Database of Suriname and Curacao). The thesis describes training Transkribus, applying the new models to scans of civil registry data, extracting information from the HTR data with regular expressions and evaluating the results. Hoek concludes that HTR quality has a large effect on extracted entity quality and suggests methods for improving these.

There are not many alternatives for Transkribus. Van der Zant, Schomaker and Haak \cite{vanderzant2008} describe MONK, an early system for hand-written text recognition developed since 2007. It is unclear what the current state of the software is\footnote{MONK website: \url{https://www.ai.rug.nl/~lambert/Monk-collections-english.html}}. Loghi\footnote{Loghi software website: https://github.com/knaw-huc/loghi} is state-of-the-art HTR software used in the Globalise\footnote{Globalise project website: \url{https://globalise.huygens.knaw.nl/project_overview/}} project for processing millions of archival pages of text. The program is freely available$^{20}$ but is not easy to use (see section \ref{sec-loghi}).

\section{Concluding remarks}

We described the project REE-HDSC and outlined our efforts to improve the quality of named entities extracted automatically from texts generated by hand-written text recognition (HTR) software. We presented a six-step processing pipeline, which we tested by processing 19th and 20th century death certificates from the civil registry of Cura\c{c}ao. We find that the pipeline extracts dates with high precision (90\%) but that the precision of person name extraction is low (33\%). Next we show how name extraction precision can be improved by retraining  HTR models with names, post-processing and by identifying and removing incorrect names.

The 2024 quality of hand-written text technology does not match the quality needs of the HDSC project. However HTR could benefit the project. In particular we are interested in getting to know if the human volunteers could speed up the annotation process if they had access to an HTR analysis of the certificates, if possible enriched with an automatic entity and date analysis. The latter could be obtained with machine learning which could be applied to different datasets without needing to be updated.

\section{Future work}

Here are our suggestions for future work, after the REE-HDSC project:

\begin{enumerate}
    \item The current number of 300 certificates (42K words) for training HTR models may be too small to create models which deliver the document quality needed for the project. We suggest to aim for at least 1000 training certificates (140K words). (Preparation for this target has already started)
    \item Both our work and the work by Hoek \cite{hoek2023} found problems with the standard reading order of lines in the Transkribus output. We suggest to develop a module which rearranges the line order in Transkribus output files. (Code for this task has already been developed by Hoek and ourselves)
    \item Unknown is yet how the HDSC volunteers will respond on having to work with computer-recognized texts. Experiments should be set up to test their response. (This task has already been picked up by the HDSC colleagues)
    \item A pipeline should be designed and tested in which a combination of computer-generated text and volunteer work is used to rapidly convert scans to data. (This task has already been picked up by the HDSC colleagues)
    \item Although the first results from the new Loghi HTR software are not encouraging, new versions may outperform Transkribus. It would be good to regularly evaluate Loghi ans compare its output text quality with that of Transkribus.
\end{enumerate}

We look forward to the future developments in the HDSC project!

\bibliographystyle{plain}
\bibliography{ref}

\appendix

\section{Instructions used in Transkribus}

This section contains lists of instructions which were used in Transkribus to process documents. We used the Transkribus Expert Client version v1.25.0 (22\_03\_2023\_07:10). Before being able to use the software the user first has to login with an account of the Transkribus website \url{https://readcoop.eu/transkribus}. There also is a web version of this software which operates differently.

\subsection{Loading a data set}
\label{sec-loading-data}

\begin{enumerate}
    \item In the top text menu of Transkribus (Server .. Tools), click on "Server"
    \item Select a collection by clicking on "Collections", select a collection from the popup menu and press "OK"
    \item Select a data set by double clicking on its name in collection name list
\end{enumerate}

We worked with the collections 168994 (Overlijdensaktes Training Curacao), 180882 (e.tjongkimsang@ esciencecenter,nl Collection) and 232117 (Overlijdensakten Cura\c{c}ao 1831-1905). Users can only see and use collections for which they were granted access by the collection owners.

\subsection{Running layout analysis and baseline detection with P2PaLA}
\label{sec-layout-p2pala}

Transkribus has two methods for layout analysis. We mainly used P2PaLA, an alternative layout analysis model available in the Transkribus Expert Client.

\begin{enumerate}
    \item Select a document collection to be processed (see appendix \ref{sec-loading-data}) 
    \item In the top text menu of Transkribus (Server .. Tools), click on "Tools"
    \item Click on "P2PaLA structure analysis tool" at the bottom (only part of the button text is visible)
    \item In the popup window select the pages to be processed: current page, all pages, or a selection
    \item Select the processing model next to "Select a model for recognition"
    \item The model filter can be used to access other models, if necessary
    \item Press "Run"
    \item Press "Yes" in the next confirmation window
\end{enumerate}

\subsection{Running standard layout analysis and baseline detection}

This section describes how to use the standard layout analysis in Transkribus. The alternative method P2PaLA is described in the previous section.

\begin{enumerate}
    \item Select a document collection to be processed (see appendix \ref{sec-loading-data}) 
    \item In the top text menu of Transkribus (Server .. Tools), click on "Tools"
    \item Under "Layout Analysis", click on "configure" to select the layout model
    \item Select the layout model next to "Selected model"
    \item Click on "OK", the selected model becomes visible above the green "Run" button
    \item Click on the highest green "Run" button
    \item Press "Yes" in the next confirmation window
\end{enumerate}

\subsection{Training layout analysis and baseline detection with P2PaLA}

Transkribus has two methods for layout analysis. We mainly used P2PaLA, an alternative layout analysis model available in the Transkribus Expert Client.

\begin{enumerate}
    \item In the top text menu of Transkribus (Server .. Tools), click on "Tools"
    \item Click on "P2PaLA structure analysis tool" at the bottom (only part of the button text is visible)
    \item In the new popup window click on "Train"
    \item In the new popup window enter a name for the new model and a description
    \item Next to "Structures", click on the green "+" and the on "OK" (selecting "paragraph")
    \item Next to "Training mode", select "Regions and Lines"
    \item Select the documents to be used for training next to "Training set"
    \item Press "Train"
    \item Press "Yes"
\end{enumerate}

\subsection{Training standard layout analysis and baseline detection}

This section describes how to train the standard layout analysis in Transkribus. Training the alternative method P2PaLA is described in the previous section.

\begin{enumerate}
    \item In the top text menu of Transkribus (Server .. Tools), click on "Tools"
    \item Under "Model Training", click on "Train a new model..."
    \item Next to "PyLaia HTR" click on "Baseline" to enable training of layout and baselines
    \item Enter a name for the new model and a description in the appropriate fields
    \item select the option "10\% from training"
    \item Select the documents to be used for training under "Documents"
    \item Press the green "+ Training" button to add the documents to the training and validation sets
    \item Click on the green "Train" button
    \item Press "Start Training" in the next confirmation window
\end{enumerate}

\subsection{Running Hand-written Text Recognition}

\begin{enumerate}
    \item Select a document collection to be processed (see appendix \ref{sec-loading-data}) 
    \item In the top text menu of Transkribus (Server .. Tools), click on "Tools"
    \item Next to "Method" under "Text Recognition", select "HTR (All engines)"
    \item Click on the green "Run" under "Method"
    \item In the popup window select the pages to be processed: current page, all pages, or a selection
    \item Make a choice between "Compute line polygons" and "Use existing line polygons" (we chose the latter)
    \item Press "Select HTR Model"
    \item Select a HTR model from the next window (not all available models may be visible)
    \item Select "Language model from training data" at the top right under "Language Model"
    \item Press "OK" to select the HTR model
    \item The selected model will be shown above the "Select HTR Model..." button, please ignore the shown recently used models
    \item Press "OK"
    \item Press "Yes" in the next confirmation window to start hand-written text recognition
\end{enumerate}

\section{Software}

All software developed in the REE-HDSC project is available on the website Github: \url{https://github.com/ree-hdsc/ree-hdsc} . This involves mostly Jupyter notebooks for data analysis but also supporting Python code and code for an interactive website for name annotation\footnote{\url{https://ifarm.nl/cgi-bin/hdsc/stats}} . Here is a list of the notebooks which were used for the various data analyses described in this report:

\begin{itemize}
    \item {\tt process\_scans.ipyb} was used for data analysis and cleaning (Section \ref{sec-data}), including drawing Figure \ref{fig-scans-per-year}
    \item {\tt layout\_analysis.ipynb} was used for layout analysis, including drawing Figures \ref{fig-layout-detection} and \ref{fig-layout-detection-improved}
    \item {\tt evaluate\_htr.ipynb} was used for obtaining the ChatGPT test results of Table \ref{tab-entity-recognition}
    \item {\tt entity\_linking.ipynb} was used for the entity linking analysis (Section \ref{sec-entity-linking})
    \item {\tt get\_names\_from\_scans.ipynb} was used for Section \ref{sec-processing-name-snippets} and Table \ref{tab-htr-improvement} about improving name recognition
    \item {\tt cer\_tests.ipynb} was used for computing the numbers in Tables \ref{tab-loghi-region} and \ref{tab-retrained}
\end{itemize}

Other notable notebooks of which the data analysis was not included in this report: 

\begin{itemize}
    \item {\tt annotator\_assessment.ipynb} computes annotator accuracies per form field 
    \item {\tt check\_printed\_text.ipynb} checks the printed text in a certificate (rather than the hand-written text) and suggests corrections
    \item {\tt htr\_training.ipynb} was used for linking unknown train file scan names to archive file names
    \item {\tt openai.ipynb} was used to communicate with ChatGPT 3.5
\end{itemize}

\pagebreak
\section{Missing districts in the deceased data of Cura\c cao}
\label{sec-missing-districts}

\begin{table}[ht]
\begin{center}
    \begin{tabular}{lcp{13cm}}
    {\bf year} & {\bf count} & {\bf missing districts} \\
    1838 & 1 & O.R. 1839 Buiten West 3e distr \\
    1842 & 1 & O.R. 1842 Buiten 9e distr \\
    1843 & 2 & O.R. 1844 Buiten 8e distr, O.R. 1844 Buiten 9e distr \\
    1845 & 1 & O.R. 1844 Buiten 9e distr \\
    1853 & 1 & O.R. 1852 Buiten 9e distr \\
    1857 & 2 & O.R. 1852 Buiten 8e distr, O.R. 1852 Buiten 9e distr \\
    1870 & 4 & O.R. 1872 Buiten 2e distr, O.R. 1872 Buiten 3e distr, O.R. 1872 Buiten 4e distr, O.R. 1872 Buiten 5e distr \\
    1871 & 1 & O.R. 1872 Buiten 5e distr\\
    1904 & 1 & O.R. 1899 Buiten 5e distr \\
    {\bf year} & {\bf count} & {\bf extra districts} \\
    1852 & 1 & O.R. 1852 Buiten 10e distr \\
    1887 & 4 & O.R. 1872 Buiten 2e distr2, O.R. 1872 Buiten 3e distr2, O.R. 1872 Buiten 4e distr2, O.R. 1872 Buiten 5e distr2 \\
    1900 & 1 & O.R. 1902 Buiten 5e distr \\
    1901 & 1 & O.R. 1902 Buiten 5e distr \\
    1912 & 1 & O.R. 1899 Buiten 6e distr \\
    \end{tabular}
\end{center}
\caption{Suspected missing districts and extra districts in the deceased data of the civil registry of Cura\c cao in 1831-1850}
\end{table}

\end{document}